# Segmentação de imagens utilizando competição e cooperação entre partículas

Silva, Bárbara Ribeiro
Breve, Fabricio Aparecido
UNESP – Universidade Estadual Paulista "Júlio de Mesquita Filho"
Rio Claro, Brasil
barbararibs@gmail.com, fabricio@rc.unesp.br

***RESUMO***: Este artigo apresenta uma proposta de extensão do modelo de aprendizado semi-supervisionado conhecido como Competição e Cooperação entre Partículas para a realização de tarefas de segmentação de imagens. Resultados preliminares mostram que esta é uma abordagem promissora.

**PALAVRAS-CHAVE:** Segmentação de imagem; Aprendizado de máquina; Aprendizado semi-supervisionado.

***ABSTRACT: This paper presents an extension proposal of the semi-supervised learning method known as Particle Competition and Cooperation for carrying out tasks of image segmentation. Preliminary results show that this is a promising approach.***

I. INTRODUÇÃO

A segmentação de imagens consiste no processo de particionamento de uma imagem em múltiplos segmentos, regiões e objetos, simplificando ou ainda mudando a forma de representação da imagem, com o objetivo de torna-la mais significativa, localizando áreas ou objetos para facilitar sua análise. Quando se trabalha com segmentação de imagens, faz-se necessário realizar uma análise detalhada, identificando oscilações nas cores, intensidade e brilho. Com o intuito de alcançar uma boa classificação é necessário entender as possíveis divisões existentes na imagem. [25]

O modelo de competição e cooperação entre partículas utiliza uma pequena porção de dados rotulados em meio aos dados não rotulados. Ao associar este modelo à tarefa da segmentação de imagens, é esperado que seja possível analisar e destacar partes ou objetos de uma imagem a partir de pequenas marcações feitas pelo usuário para direcionar o algoritmo.

Existem diversas formas de segmentação, portanto é necessário analisar o confinamento dos grupos que cada algoritmo proporciona e os aspectos gerais extraídos de acordo com características da imagem e capacidade de processamento da máquina onde o algoritmo escolhido será executado.

Utilizado no modelo de competição e cooperação entre partículas, o aprendizado de máquina é caracterizado por algoritmos que melhoram automaticamente com a experiência, na tentativa de imitar o comportamento de aprendizado humano. Tal comportamento pode ser extraído com ajustes de parâmetros baseando-se nos dados de entrada apresentados ou ainda baseados nas informações de saída esperadas.

Aprendizado semi-supervisionado é uma das categorias de aprendizado de máquina que tem recebido bastante atenção nos últimos anos. Nela dados rotulados e não rotulados são combinados a fim de se obter melhor classificação. É diferente do que acontece no aprendizado supervisionado, onde apenas dados rotulados são utilizados, e do aprendizado não-supervisionado, onde informações de rótulos não são fornecidos.

Algumas técnicas de aprendizado semi-supervisionado foram aplicadas com sucesso em tarefas de classificação de imagens. Nesses casos, as informações de rótulos de alguns pixels da imagem são fornecidas pelo usuário, e o algoritmo deve ser capaz de propagar os rótulos para os demais pixels da imagem. Um exemplo são as técnicas baseadas em redes neurais artificiais e as técnicas baseadas em grafos. [22][13]

O modelo de competição e cooperação entre partículas [10] é uma abordagem de aprendizado semi-supervisionado recente, que faz o uso de partículas que caminham em um grafo, seguindo os mecanismos de competição e cooperação com o intuito de demarcar territórios e propagar rótulos.

Essa abordagem se destaca por apresentar uma propagação local de rótulos, feita através das partículas, em vez de uma propagação global como acontece na maioria dos métodos baseados em grafos. Isto lhe proporciona um menor custo computacional, tornando-o adequado para aplicação em bases de dados maiores.

A abordagem de competição e cooperação, apesar de recente, já foi utilizada em diversas tarefas como classificação de dados [10], detecção de comunidades sobrepostas [7] [9], aprendizado com dados imperfeitos [6], aprendizado ativo [4], aprendizado com mudança de conceitos [5], etc. Porém sua aplicação sempre foi feita em dados baseados em vetores de atributos, nunca em imagens. A extensão do modelo para aplicação na tarefa de segmentação de imagens não é uma função trivial, porém é bastante promissora.

No algoritmo de competição e cooperação entre partículas original, os dados vetoriais são transformados em uma rede complexa com base na distância Euclidiana entre os pares de amostras. No caso de imagens, as redes são geradas com base não apenas nas similaridades e diferenças entre características extraídas dos pixels, mas também nas distâncias espaciais entre eles na imagem original.

A segmentação autônoma representa uma das tarefas mais complexas quando trabalhamos com processamento de imagens [33]. Para simplificar esta tarefa utilizamos nesta proposta o modelo de competição e cooperação entre partículas, que utiliza aprendizado semi-supervisionado. Desta forma, é possível fazer uso de um especialista para o auxílio na segmentação de imagens, rotulando manualmente alguns pixels de cada objeto que se deseja destacar. O algoritmo se encarregará de completar a tarefa, usando para isso ambos os pixels rotulados e não rotulados como fonte de aprendizado.

O objetivo desta proposta é estender o modelo de competição e cooperação entre partículas para tratar o problema de segmentação de imagens. Deste modo, partículas serão criadas através dos pixels da imagem identificados pelo usuário e, assim tais rótulos poderão se espalhar para os demais pixels.

No decorrer do artigo serão expostos os principais conceitos das técnicas utilizadas, como segmentação de imagens, aprendizado de máquina com ênfase no aprendizado semi-supervisionado e o modelo de competição e cooperação entre partículas. Situando sobre as temáticas relacionadas, serão expostas as análises e formas em que foram unidos tais conceitos.

II. CONCEITOS E TÉCNICAS

*A. Segmentação de Imagens*

Existem diversas formas de extrair informações de uma determinada imagem. Uma destas técnicas consiste na transformação da imagem em um cluster de pixels, para serem estudados individualmente. Dessa forma previnem-se perdas de informações.

Para a escolha do método de segmentação é necessário analisar o confinamento de grupos que o algoritmo proporciona e tipos de aspectos gerais que serão extraídos [19]. Tais agrupamentos podem corresponder aos aspectos globais da imagem. No entanto é necessário analisar o tempo e critério de execução por linha de pixels, obtendo um método prático e eficiente.

As imagens mais complexas de segmentar são as coloridas. Para este tipo de imagem é necessário utilizar um processo de extração de regiões que possuam homogeneidade, podendo ser analisadas propriedades geométricas ou tons [26]. É possível fazer uma análise com base nos valores RGB, estabelecendo uma medida de distância Euclidiana entre valores de pixels, e assim medir a distância entre os tons. [17]

Na fase de processamento da imagem é necessário levar em conta o nível de processamento que deverá ser utilizado; para tal tarefa é aplicado o paradigma de divisão do processamento da imagem em três tipos [16]. O processamento primitivo para a redução do ruído da imagem ocorre no processamento de baixo nível, denominado desta forma devido à entrada e a saída serem imagens. A segmentação entra no processo de nível médio do processamento, devido ao fato da entrada ser uma imagem e os resultados da saída serem atributos extraídos da imagem. O processamento de alto nível corresponde à busca por conjuntos de objetos ou partes que sejam reconhecidas e estão associadas na imagem, logo a entrada e a saída são atributos extraídos previamente da imagem [17].

A escolha de um determinado algoritmo é dada pela avaliação da imagem. É necessário que, no retorno da segmentação, a imagem apresente um nível de detalhamento intermediário. Não é necessário um nível de detalhamento muito alto, porém deve-se evitar um detalhamento muito grosseiro [20].

A abordagem de segmentação utilizada nesta proposta é baseada em grafos. Neste tipo de abordagem cada pixel da imagem é representado como um nó do grafo, de forma que a presença ou falta de vizinhança determina a afinidade entre eles, formando arcos a fim de representar pontos com maiores afinidades [22].

O objetivo da segmentação é definir fronteiras e distinção da intensidade entre os pixels, dentro de uma determinada região. Algumas características podem ser extraídas com o conhecimento da distância entre nós e características comuns em nós próximos, sendo possível identificar dentro da imagem diferenças de intensidade do brilho, cor, entre outros. É necessário estabelecer critérios que tornem possível distanciar pixels que representem diferentes objetos dentro da imagem, bem como aproximar pixels que represente o mesmo objeto ou semelhante, mesmo que este objeto seja representado em áreas separadas por demais objetos.

Ao trabalhar com imagens coloridas é necessário incorporar alterações ao método, devido à semelhança entre cores sobrepostas. A solução para tal problema é baseada na montagem de um grafo multidimensional. A técnica consiste na divisão dos pixels no padrão RGB (vermelho, verde e azul), em união com os pontos X e Y da imagem em conjuntos.

*B. Aprendizado de Máquina*

O aprendizado de máquina permite o desenvolvimento de algoritmos que melhoram automaticamente com a experiência. Tal comportamento pode ser obtido através do ajuste de parâmetros com base nos dados de entrada que são apresentados e, em alguns casos, nas informações de saída desejadas [21] [20] [3].

Um dos principais objetivos é aprender automaticamente a reconhecer padrões complexos além de tomar decisões inteligentes com base em dados. No entanto a dificuldade está na complexidade da descrição de formas para linguagem de programação. Este tipo de técnica tem o intuito de construir sistemas que possam se adaptar a circunstâncias específicas, sem que seja necessário escrever um novo programa para cada situação nova [21] [20] [3].

No aprendizado supervisionado, um algoritmo aprende a função de classificação a partir dos dados de treinamento, os quais consistem em pares de itens de dados e seus respectivos rótulos, de forma que após "aprender" um padrão no comportamento dos dados, o algoritmo pode prever os rótulos de novos dados. Por outro lado, no aprendizado não supervisionado, todos os itens de dados não são rotulados e o objetivo é determinar suas estruturas. O aprendizado semi-supervisionado encontra-se no meio termo entre o aprendizado supervisionado e o aprendizado não supervisionado.

*C. Aprendizado Semi-Supervisionado*

O aprendizado semi-supervisionado foi criado com o intuito de proporcionar maior praticidade, desempenho e baixo custo. [33] Com o crescimento constante da oferta de dados online, é fácil encontrar dados de todos os tipos, porém em sua maioria, estes dados não estão classificados ou são poucos os que estão. A partir deste cenário torna-se difícil classificar todos ou a maioria dos dados manualmente. Para solucionar tal empecilho é possível utilizar os dados classificados e/ou classificar apenas uma parte pequena e tomar estes dados como exemplo para a classificação dos demais dados da base.

Este tipo de aprendizado se aproxima do modo que os seres humanos reconhecem e aprendem o que ocorre a sua volta, tendo em vista que muitas situações são desconhecidas durante um período, porém depois de analisadas e comparadas com outras situações acabam por se tornar comuns e simples.

A implementação de tal técnica utiliza algoritmos que trabalhem com a comparação entre pares de dados, onde a classificação ocorre com base na proximidade e/ou distância entre os dados comparados. Pode-se trabalhar com clusters, onde dados que estejam no mesmo grupo estão suscetíveis a pertencerem à mesma classe. Também podem apresentar regiões de baixa densidade, onde as classes, apesar de separadas, não são fortemente espaçadas. Em contraposição pode haver regiões de alta densidade, onde as classes ficam nitidamente distintas [3].

Nestes casos, métodos de aprendizado semi-supervisionado se tornam interessantes. Eles tratam este problema específico combinando poucos itens de dados rotulados com uma grande quantidade de dados não rotulados para produzir melhores classificadores, ao mesmo tempo em que requerem menor esforço humano [1] [12] [32].

A categoria mais ativa recentemente é a de técnicas baseadas em grafos. Nesta categoria os dados são representados por nós de um grafo, e as arestas que interligam estes nós indicam a similaridade entre os mesmos.

Tais métodos assumem que existe suavidade dos rótulos no grafo. A maioria dos métodos dessa categoria podem ser vistos como um framework de regularização que estima uma função *f* que satisfaça tanto a suavidade dos rótulos quanto a proximidade dos rótulos estimados aos dados pré-rotulados [35] [31] [30] [28] [27] [18] [34] [29].

Dentro de um grafo os nós consistem na marcação dos pontos de dados, onde as arestas representam as semelhanças entre pontos. As arestas são utilizadas como caminho para que os rótulos se propaguem através do grafo de maneira a tentar descobrir os rótulos desconhecidos. Os métodos que se baseiam em grafo permitem a conversão dos dados em uma matriz, onde o objetivo é a minimização de custo da classificação geral. O critério de custo é introduzido quando ocorrem mudanças suaves nos dados, onde os dados mais próximos de uma mesma classificação tendem a ter como vizinhos próximos dados da mesma classe [19].

Quando um grafo está sendo analisado, e são encontrados dois nós cuja aresta que os ligam possui um peso baixo, a tendência é que estes sejam nós pertencentes a diferentes classes, ou classes iguais mas com características menos homogêneas.

Para a extração de conteúdo do grafo podemos utilizar diversas técnicas, como passeios aleatórios [18], campos randômicos gaussianos[27], consistência global e local [31], e diversos outros métodos.

O modelo de competição e cooperação entre partículas, utilizado nesta proposta, se enquadra na categoria dos métodos baseados em grafos. Porém sua abordagem é diferente da utilizada pelos demais métodos dessa categoria, como será visto na próxima seção.

### III. METODOLOGIA DE DESENVOLVIMENTO

Nesta proposta é utilizado o modelo de competição e cooperação entre partículas, que utiliza aprendizado semi-supervisionado, de modo que um especialista possa auxiliar no processo de segmentação, rotulando manualmente alguns pixels de cada objeto da imagem. Desta forma o algoritmo se encarregará de propagar os rótulos dados pelos especialistas para pixels não rotulados com base nas informações extraídas de outros pixels sejam eles rotulados ou não.

Uma primeira abordagem de competição de partículas foi desenvolvida para detectar comunidades em redes [24]. Neste modelo, partículas caminham em uma rede e competem umas com as outras tentando possuir a maior quantidade possível de nós de um grafo. Ao mesmo tempo, cada partícula evita que outras partículas invadam seu território. Finalmente, cada partícula permanece confinada dentro de uma comunidade da rede.

Com base na abordagem de competição de partículas, [10] e [8] desenvolveram um modelo de aprendizado semi-supervisionado baseado em grafos. Dentre várias melhorias, o modelo novo apresenta mecanismos de competição e cooperação combinados em um esquema único. Partículas representando uma mesma classe caminham em uma rede de maneira cooperativa para propagar seu rótulo. Ao mesmo tempo, partículas de diferentes classes competem entre si para determinar a borda das classes. Este algoritmo é capaz de classificar dados linearmente inseparáveis e com um tempo de execução menor que outros algoritmos baseados em grafo tradicionais, devido à menor complexidade computacional. O modelo de partículas tem complexidade computacional apenas linear (*O(n)*, onde *n* é a quantidade de nós do grafo) no laço principal [10], enquanto a maioria dos demais modelos baseados em grafo tem complexidade cúbica [32].

Nos modelos tradicionais de aprendizado semi-supervisionado baseado em grafos, a informação de rótulos é propagada de todos os nós para todos os nós em cada passo do algoritmo, levando em consideração os pesos das arestas, como por exemplo no modelo apresentado por [31]. No modelo de partículas a propagação de rótulos ocorre de forma local, ou seja, em cada passo do algoritmo cada partícula escolhe um vizinho para propagar seu rótulo. Deste modo, cada partícula visita apenas nós que potencialmente pertençam ao seu time, não visitando nós que já estejam dominados por outra partícula.

Realizada a leitura da imagem, é necessário montar a estrutura do grafo para que o modelo de movimentação de partículas possa ser aplicado. Para a montagem do grafo é utilizado o conjunto de dados $X = x_1, x_2, ..., x_n \subset R^m$ a ser analisado e transformado em uma rede sem pesos e não direcionada, e um conjunto de rótulos $L = 1, 2, ..., c$ onde estes rótulos são escolhidos para que sejam anexadas marcações de cada um dos times.

Com tais informações é montada uma rede, representada através de um grafo $G = (V, E)$ com $V = v_1, v_2, \ldots v_n$ onde cada nó $v_i$ corresponde a um pixel da imagem. A matriz de adjacência $W$ define quais nós da rede são interconectados:

$$W_{ij} = 1 \; se \; \|x_i - x_j\|^2 \leq \sigma \; e \; i \neq j$$
$$W_{ij} = 0 \; se \; \|x_i - x_j\|^2 \leq \sigma \; e \; i = j \quad (1)$$

onde $W_{ij}$ especifica se há aresta entre os nós $x_i$ e $x_j$ e $\sigma$ é um limiar de distância.

Outra forma de construir a matriz de adjacências é verificando $x_i$ está entre os $k$ vizinhos mais próximos de $x_j$, ou vice versa. No caso verdadeiro $W_{ij} = 1$ e $W_{ij} = 0$ caso contrário.

Para cada pixel rotulado, além de um nó, também é gerada uma partícula. O conjunto total de partículas é dado por $P = (p_1, p_2, \ldots, p_c)$. Cada partícula $p_i$ possui uma variável que representa o potencial da partícula e é dado por:

$$p_j^w(t) \in [w_{min} \; w_{max}] \quad (2)$$

onde $w_{min} = 0$ e $w_{max} = 1$. A posição inicial de cada partícula é o nó rotulado à que ela corresponde, chamado "nó casa" da partícula.

Conforme as partículas mudam de posição, a distância do nó atual para o nó casa é registrada. Partículas geradas de amostras da mesma classe agem como um time, colaborando com as outras e competindo com partículas de outros times.

Cada nó da rede tem um vetor

$$v_i^w(t) = v_1^{w1}(t), \quad v_2^{w2}(t), \ldots, v_i^{wL}(t) \quad (3)$$

em que cada elemento $v_i^{wj}(t) \in [\omega_{min} \; \omega_{max}]$ representa o nível de dominância do time $l$ sobre esse nó $v_i$. A soma desse vetor é sempre constante, $\sum_{l=1}^{c} v_i^{\omega l} = 1$, devido ao fato de que quando uma partícula aumenta o nível de domínio de seu time elas diminuem proporcionalmente o nível de domínio dos demais times.

Outra variável que cada partícula recebe é um vetor de distâncias

$$p_j^d(t) = p_j^{d1}(t), p_j^{d2}(t), \ldots, \quad p_j^{dn}(t) \quad (4)$$

onde cada elemento $p_j^{di}(t) \in [0, n-1]$ equivale à distância medida entre o nó-casa da partícula $p_j$ e o nó $v_i$.

No decorrer da execução do programa, as partículas exercem sua caminhada pela rede para aumentar o nível de dominância de seu time sobre o nó, ao mesmo tempo em que diminuem o nível de dominância dos outros times. Cada partícula possui um nível de força, que aumenta quando ela visita nós dominados por seu time, e diminui quando ela visita um nó dominado por outro time. Essa força é importante porque a mudança que a partícula causa em um nó é proporcional à força que ela possui naquele momento.

A cada iteração cada partícula seleciona um nó vizinhos a ser visitado. Nessa ocasião a partícula eleva o domínio de seu time/classe nesse nó e ganha ou perde força de acordo com o nível de domínio resultante. Em seguida ela verifica qual é o time dominante naquele nó, através do vetor de domínio. Caso a classe da partícula seja a mais forte no nó, a partícula permanece no nó que está sendo visitado. Caso contrário ela volta ao nó em que estava anteriormente.

Este mecanismo garante que a partícula fique mais forte quando está em sua vizinhança, protegendo-a; e fique mais fraca quando está tentando invadir outros territórios. Inicialmente, todas as partículas têm sua força configurada no nível máximo e todos os nós rotulados têm nível de dominância configurado no máximo para o time correspondente, enquanto os demais nós têm nível de dominância distribuído igualmente entre os times:

$$v_i^{wl} = 1 \; se \; y_i = l,$$
$$v_i^{wl} = 0 \; se \; y_i \neq l \; e \; y_i \in l, \quad (5)$$
$$v_i^{wl} = \frac{1}{c} \; se \; y_i = 0.$$

A tabela de distância introduzida em cada partícula tem como objetivo mantê-las com um controle da distância entre o "nó atual" e o "nó casa", evitando assim que as partículas de distanciem muito, deixando suas vizinhanças e ficando suscetíveis a ataques de partículas pertencentes a outras classes. Ao escolher um nó a ser visitado, as partículas levam em consideração essa distância, juntamente com o nível de domínio de seu time em cada vizinho.

Para calcular a distância entre a partícula e cada nó inserido no grafo, assume-se inicialmente um conhecimento limitado da rede, no qual a partícula sabe apenas que no grafo há $n$ nós. No entanto, ela não sabe exatamente a distância entre o seu "nó casa" e cada um deles, assim a partícula inicializa a distância de todos os nós como $n - 1$. A cada visita a partícula atualiza a informação de distância que tem até o nó em questão dinamicamente.

As partículas escolhem os nós a serem visitados com base em uma de duas regras. A cada iteração a partícula escolhe uma das regras aleatoriamente com probabilidades pré-definidas. As duas regras são descritas a seguir:

- Regra aleatória: a partícula escolhe aleatoriamente, com iguais probabilidades, qualquer um dos nós vizinhos do nó em que ela se encontra. A regra aleatória não se preocupa com os níveis de domínio ou distância do nó-casa com os possíveis nós a serem visitados, sendo útil para exploração e aquisição de novos nós.

- Regra gulosa: a partícula escolhe aleatoriamente qualquer um dos nós vizinhos do nó em que ela se encontra, com probabilidades calculadas de forma diretamente proporcionais ao nível de dominância do time dessa partícula em cada nó, e inversamente proporcionais à distância de cada nó ao nó casa dessa partícula.

Dessa forma, ao escolher a regra gulosa, as partículas tendem a ficarem em territórios já dominados e preferem nós próximos de sua origem (em sua vizinhança), ou seja, elas assumem um comportamento mais conservador e defensivo. Por outro lado, ao escolher a regra aleatória as partículas ignoram níveis de dominância e distâncias e tem maior probabilidade de escolher nós que não dominam e nós distantes de sua origem, assumindo assim um comportamento de exploração maior.

IV. RESULTADOS PRELIMINARES

Utilizando o algoritmo descrito e exposto neste artigo é possível obter uma classificação considerável principalmente utilizando imagens simples e com poucos detalhes.

Para calcular a distância Euclidiana entre os pares de pixels foram utilizadas as posições *x* e *y* de cada pixel, em relação a imagem original, em união com os valores RGB de cada pixel.

Para ilustrar o funcionamento do algoritmo serão utilizados dois tipos de imagens, a Figura 1 representa uma imagem simples de fácil análise para o algoritmo, e a Figura 4, já com as marcações para ser processada pelo algoritmo, a qual representa a atuação do algoritmo diante de uma imagem real, com mais detalhes.

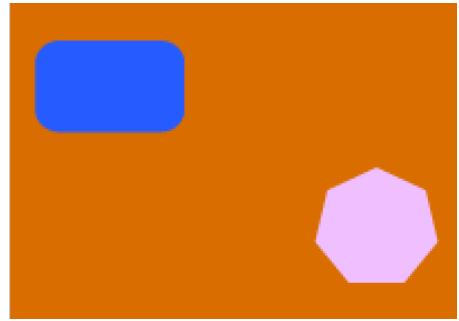

Figura 1 – Imagem original sintética com 200x140 pixels.

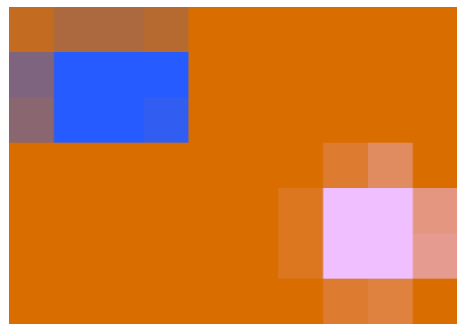

Figura 2 – Imagem sintética redimensionada para 10X7 Pixels.

A Figura 1, inicialmente com 200X140 pixels, foi reduzida para 10X7 pixels, Figura 2. Esta alteração é feita para permitir a representação gráfica na Figura 7, ilustrando o grafo formado a partir da imagem, para melhor compreensão do funcionamento do algoritmo.

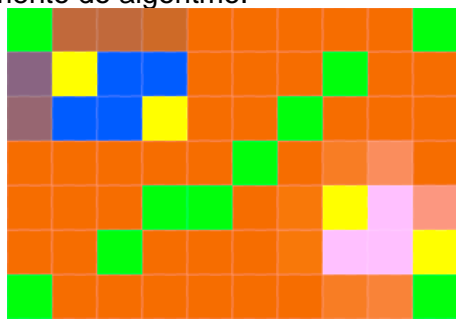

Figura 3 – Imagem sintética incluindo as marcações feitas pelo usuário.

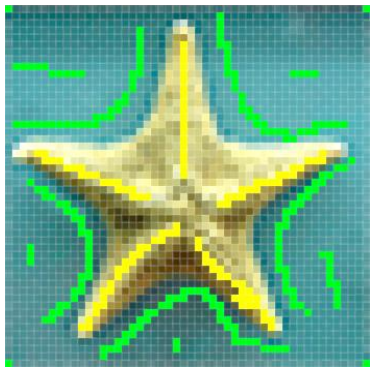

Figura 4 - Imagem real de entrada, incluindo as marcações feitas pelo usuário.

Cada imagem de entrada é formada por duas camadas. A primeira camada, mostrada nas Figuras 2 e 5, armazena a imagem original, que será utilizada para montar o grafo. A segunda camada, ilustrada na Figura 6, contém as informações de quais pixels foram rotulados pelo especialista e seus respectivos rótulos. Esta camada é utilizada para definir o conjunto de partículas e as informações iniciais de domínio dos nós.

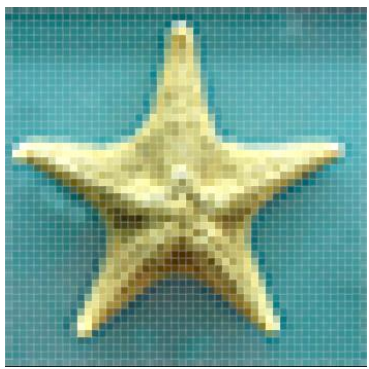

Figura 5 – Primeira camada da imagem da Figura 4, representando a imagem original.

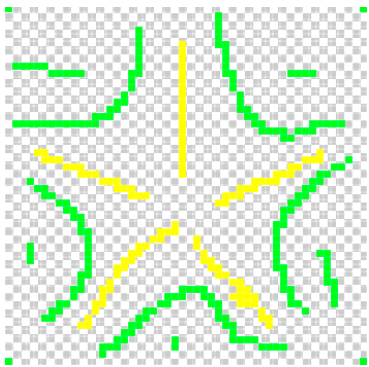

Figura 6 – Segunda camada da imagem da Figura 4, representando as marcações feitas pelo usuário.

A Figura 7 mostra o grafo montado a partir da Figura 3, utilizando $k = 10$. Observando este grafo é possível perceber e entender melhor a distribuição e ligação dos pixels. Os números em cada nó equivalem ao número do pixel dentro da matriz preenchida em tempo de execução.

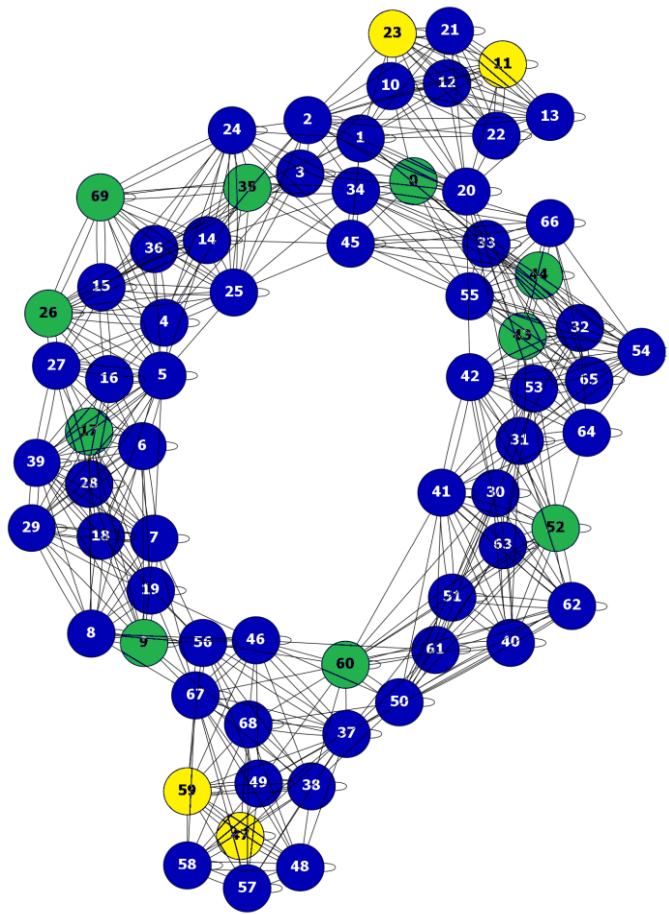

Figura 7 – Grafo formado a partir da imagem da Figura 3

O resultado final fornecido pelo programa equivale a demarcação dos pixels com as cores correspondentes às partículas que conquistaram aquela posição. O resultado da Figura 8 foi obtido ao receber como entrada a Figura 3, assim como a Figura 9 foi obtida ao receber a Figura 4 como entrada.

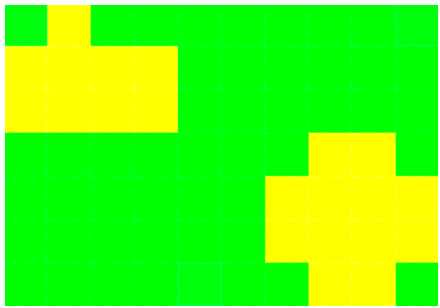

Figura 8 – Classificação baseada na Figura 3

Apesar da Figura 9 apresentar pequenos erros de classificação, é possível reconhecer a representação original. As cores utilizadas para o desenho são baseadas nas cores das partículas, representando o pixel rotulado e pertencente ao grupo da partícula.

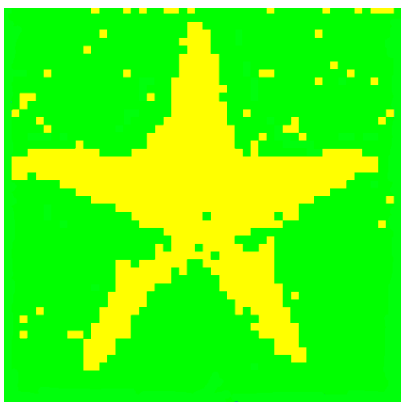

Figura 9 – Resultado da classificação baseada na Figura 4

As Figuras utilizadas até o momento foram escolhidas com o objetivo de testar o funcionamento do modelo de movimentação de partículas, e as diversas formas de se trabalhar com os componentes RGB e coordenadas de posicionamento dos pixels. Posteriormente o algoritmo será melhorado para que possa receber imagens reais mais complexas.

## V. CONCLUSÃO

Neste artigo é apresentada uma extensão do modelo de competição e cooperação entre partículas para tratar a tarefa de segmentação de imagens.

Os resultados das simulações computacionais com imagens simples mostram que tal abordagem é bastante promissora. O algoritmo vem sendo aperfeiçoado com o intuito de melhorar a classificação de imagens simples e possibilitar também a classificação de imagens mais complexas, mais ricas em detalhes e cores.

Como trabalho futuro, também pretende-se permitir que rótulos sejam inseridos ou apagados pelo especialista em tempo de execução, de forma interativa. Isto exigirá que o algoritmo seja capaz de se adaptar a inserção e remoção de partículas em tempo de execução.

## VI. REFERÊNCIAS

**Barbara Ribeiro da Silva** recebeu seu diploma de bacharel da Universidade Estadual "Júlio de Mesquita Filho", Brasil em 2012. Atualmente Mestranda da Universidade Estadual "Júlio de Mesquita Filho"- UNESP. Seus interesses de pesquisa incluem a aprendizagem de máquina, processamento de imagens, segmentação de imagens e inspirados na natureza computação.

**Fabricio Aparecido Breve** possui graduação em Ciência da Computação pela Universidade Metodista de Piracicaba (2001), mestrado em Ciência da Computação pela Universidade Federal de São Carlos (2006) e doutorado em Ciências da Computação e Matemática Computacional pela Universidade de São Paulo (2010), com período sanduíche na University of Alberta, Canadá. Atualmente é professor assistente doutor da Universidade Estadual Paulista Júlio de Mesquita Filho. Seus interesses de pesquisa incluem aprendizado de máquina, reconhecimento de padrões, processamento de imagens, redes complexas, redes neurais artificiais e computação inspirada pela natureza